%% file: main.tex
\documentclass[10pt,twocolumn,letterpaper]{article}

\pdfoutput=1

\usepackage{cvpr}
\usepackage{times}
\usepackage{epsfig}
\usepackage{graphicx}
\usepackage{amsmath}
\usepackage{amssymb}
\graphicspath{{./pictures/}}
\usepackage[table]{xcolor}
\usepackage{booktabs}
\usepackage{stmaryrd}
\usepackage{wasysym}
\usepackage{pifont}
\usepackage{url}

\newcommand\tstrut{\rule{0pt}{2.4ex}}
\newcommand\bstrut{\rule[-1.0ex]{0pt}{0pt}}

\usepackage[pagebackref=true,breaklinks=true,letterpaper=true,colorlinks,bookmarks=false]{hyperref}

\cvprfinalcopy 

\ifcvprfinal\fi
\begin{document}

\title{iCub World: Friendly Robots Help Building Good Vision Data-Sets\thanks{This work was supported by the European FP7 ICT project No. 270490 (EFAA) and project No. 270273 (Xperience).}}

\author{Sean Ryan Fanello\textsuperscript{1,2}, Carlo Ciliberto\textsuperscript{1}, Matteo Santoro\textsuperscript{3}, Lorenzo Natale\textsuperscript{1}, Giorgio Metta\textsuperscript{1},\\
Lorenzo Rosasco\textsuperscript{2,3}, Francesca Odone\textsuperscript{2}\\
\textsuperscript{1}iCub Facility, Istituto Italiano di Tecnologia, Genova, Italia\\
\textsuperscript{2}DIBRIS, Universit\`{a} degli Studi di Genova, Genova, Italia\\
\textsuperscript{3}LCSL, Istituto Italiano di Tecnologia and Massachusetts Institute of Technology \\
{\tt\small \{sean.fanello,carlo.ciliberto,matteo.santoro,lorenzo.natale,giorgio.metta\}@iit.it, } \\
{\tt\small lrosasco@mit.edu, francesca.odone@unige.it }
}

\maketitle

\begin{abstract}
In this paper we present and start analyzing the  iCub World data-set, 
an object recognition data-set, we acquired using  a Human-Robot Interaction (HRI) 
scheme and  the iCub humanoid robot platform. Our set up  allows for rapid acquisition and annotation of data  with corresponding ground truth. 
While more constrained in its scopes -- the iCub world is essentially a robotics research lab -- we demonstrate  how the proposed data-set poses challenges to  current recognition systems. 
The iCubWorld data-set is publicly available \footnote{The data-set can be downloaded from: \url{http://www.iit.it/en/projects/data-sets.html}}.
%
%
%
\end{abstract}

\input{introduction}

\input{system}
\input{experiments}
\input{discussion}

{\small
\bibliographystyle{ieee}
\bibliography{biblio}
}

\end{document}

%% file: introduction.tex
\section{Introduction}

 



%
%
%
%

The availability of large data sets, e.g.  Caltech-101 \cite{feifei04}, PASCAL VOC \cite{pascal}, ImageNet \cite{imagenet}, SUN \cite{sun}, has had a major impact in computer vision.  Notably, it has enabled rapid benchmarking of different algorithms and encouraged reproducible research. 
Data-sets, such as Caltech-101 or ImageNet, are very wide and ambitious in their scopes, in that they 
aspire to represent the whole (or a large portion) of the visual world. Indeed, a variety of challenges arise in this context.  
Image content has a  semantic hierarchy and  object classes have wide variability  due to intrinsic (objects instances may differ) or extrinsic factors (illumination, view-point, occlusions and shadows). Textures and  geometry can be discriminative factors. 
Source of nuisance include objects spatial extent and scale, background vs context, the presence of multiple objects, possibly with a different focus of interest. Not surprisingly, building good data-sets with such a broad scope is tricky.  For example, data gathering (labeling) is cumbersome and, most importantly, data-sets turn out to have often strong biases which prevent generalization ~\cite{torralba}. Indeed, 
image retrieval rather than image understanding becomes often the key question.

In this work, we shift our attention to the more restricted, yet challenging, world of the iCub humanoid robot \cite{metta2010}. Visual tasks in this setting are naturally motivated by further robotics tasks, e.g. navigation~\cite{Filliat07} and manipulation~\cite{kragic03}.  
Beyond robotics, the iCub can be seen as a full-body emulation of the human complexity and the iCub world becomes a natural simplification of the world where humans live. In this sense, the perceptual challenges presented to the iCub are similar to those faced by biological beings. 

The use of  a humanoid robot for the acquisition (and annotation) of vision datasets has a natural appeal in terms of rapid data  gathering and ground truth acquisition. As discussed in the following, data acquisition and annotation is considerably simpler since it relies on a natural Human Robot Interaction (HRI) scenario. Human labeling is here replaced by vocal and gesture interaction of the human supervisor with the robot. Also, a more controlled environment allows us to reduce biases in the data while tuning the  amount of nuisance factors.

\begin{figure}[t]
\begin{center}
   \includegraphics[width=\columnwidth]{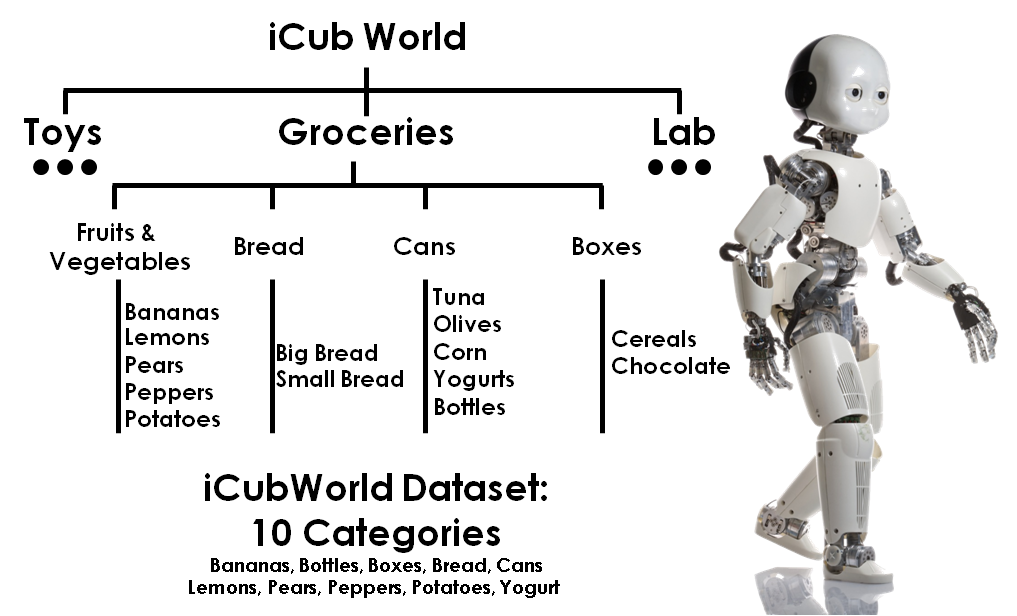}
\caption{The hierarchy of the iCub World. The robot lives in a lab environment where toys, food and office tools are mainly present. For the proposed dataset we selected the groceries subset of the hierarchy. More categories will be available in the next releases.}
\label{fig:tree}
\end{center}
\end{figure}

The \textit{iCubWorld} is an object recognition data-set we started building following the above scheme. 
It is somewhat complementary to other vision data-sets in robotics  which are acquired in considerably more 
constrained settings with a strong supervision on pose and object location during the training phase 
(e.g. RGBD Object Dataset \cite{lai2011}, Ikea Kitchen Dataset \cite{ikea}). 
The iCubWorld  data-set currently comprises 10 food  categories, see Figure \ref{fig:tree}, that we will extend  
over time for example including toys and typical lab objects. 
Although the iCubWorld is more restricted in its scopes, it retains challenging aspects of already available data-sets, 
while offering new ones caused by the physical limitations of the robot.

We demonstrate the challenges of the iCubWorld by addressing object categorization with a a representative set of 
state of the art visual recognition methods. More precisely, we consider systems based  on supervised learning machines
coupled with a two layers feature extraction/learning systems.
The first layer is based on local/low level features (SIFT) while the second layer extract  higher level features (bag of words \cite{csurka04}, sparse coding \cite{lee07, yang09}, locally-constrained linear coding \cite{wang10}).   The HMAX biologically inspired architecture is also considered \cite{serre07}.

Our results show these vision systems can achieve good performances in the iCub world, but performances can drop drastically 
as soon as the acquisition conditions change, a new object instance is considered for a known category, the object is held by a human other than the original supervisor or by the iCub itself. 

The reminder of the paper is organized as follows. In Section 2 we describe the iCub setting, in Section 3 we summarize the current state of the iCubWorld data-set. In Section 4 we summarize the state-of-the-art on visual recognition we are referring to, while Section 5 reports the results of our experimental analysis. Section 6 is left to a final discussion.

\begin{figure*}[t]
\begin{center}
   \includegraphics[width=1.0\linewidth]{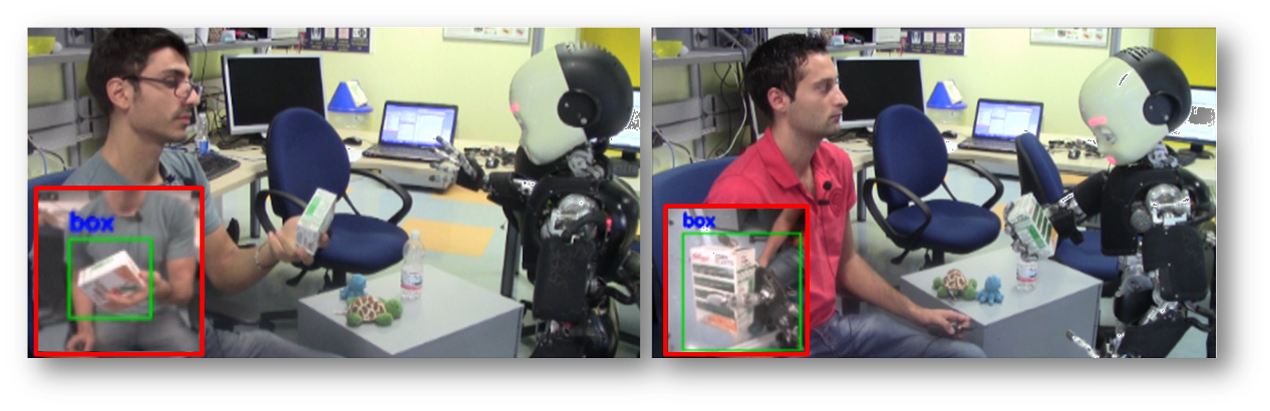}
\caption{The Human-Robot Interaction setting. The iCub's motion detector (left) and forward kinematics (right) both provide a reasonably good estimation of the object position withing the image.}
\label{fig:hri}
\end{center}
\end{figure*}

%% file: system.tex
\section{The iCub HRI setting}\label{sec:setting}

The Human Robot Interaction (HRI) setting presents challenging tasks for learning systems. Indeed, robots hold only limited knowledge about the world - the information accessible through their sensors - and yet they are typically required to generalize these partial observations to more general contexts. HRI however, is suited for non expensive acquisition of annotated data: the communication between the human and a humanoid robot is required to occur exclusively on natural channels (e.g. speech) and thus manual labeling of individual samples can be avoided. 

In the proposed scenario, the iCub robot~\footnote{The iCub is a humanoid robot with $53$ degrees of freedom, equipped with two actuated digital cameras, microphones for speech acquisition, inertial and force/torque sensors. Additionally, the torso, arms, hands and fingertips of the iCub are covered with artificial skin that provides haptic feedback. The software running on the iCub - including the proposed HRI scheme - is released under the GPL license and it is publicly available for download  \url{www.icub.org}. }  \cite{metta2010} is shown one object at a time so that it can learn the object's appearance to be able to recognize it in the future. A demonstrator provides verbal annotation, pronouncing the category name while presenting the object. 

We propose two different modalities  to exploit the contextual information and to perform the approximate localization of the objects within the images:

\begin{itemize}
\item\textbf{Human Mode (Fig.~\ref{fig:hri} - left)}. The demonstrator moves the object in front of the robot, so that the robot can observe the object from different viewpoints. An independent motion detector~\cite{motionCUT,fanello13b} is employed to identify a bounding box around the moving target and to have the robot actively track it with its gaze.

\item\textbf{Robot Mode (Fig.~\ref{fig:hri} - right)} The demonstrator gives the object to the robot that starts to move its hand in order to observe the object from multiple points of view. The system exploits the kinematic structure of the robot to track the hand (thus the object) in the images and identify a bounding box around it.
\end{itemize}

\section{The iCubWorld data-set}
\label{sec:dataset}

In Fig. \ref{fig:tree} the sketch of a tentative hierarchy of objects of interest for the iCub is depicted. In this work we selected a subset of groceries bought in a local supermarket. Further acquisitions regarding lab/office-type objects (pens, books, etc.) are currently in progress. The very final goal of the iCubWorld project is to obtain a rich data-set reproducing with good accuracy a typical domestic environment.
Following the HRI schema described in Sec.~\ref{sec:setting}, we collected the iCubWorld data-set currently comprising $10$ visual object categories. We selected objects of different complexity, shape and texture. For each category we provided $4$ different object instances and for each instance $200$ examples (examples of one instance per category are shown in Fig.~\ref{fig:dataset}).
The training set has been acquired in the \textit{Human Mode} and, to mimic a learning session where a supervisor instructs the robot to recognize new objects, a single demonstrator shows the objects to the robot. For each category only $3$ object instances used for the training. Overall, each object category includes $600$ images per category.
In the test set we include examples of all the $4$ objects instances per each class, exploit both acquisition modalities, {\em Robot} and {\em Human}, and a different human demonstrator. 
The original $640 \times 480$ pixels images acquired from the iCub cameras have been automatically cropped (see Sec.~\ref{sec:setting}), obtaining  $160 \times 160$ images for the \textit{Human} mode and $320 \times 320$ for the \textit{Robot} mode respectively. 

The proposed data-set presents two main differences with respect to most typical categorization benchmarks. First, the HRI domain allows us to design the acquisition setting so that annotation can be performed automatically. This efficiently reduces the influence of the (possibly arbitrary) interpretation of an individual supervisor. Indeed, in our case the selection bias is limited to the original choice of the category representatives, while in typical image retrieval settings each single image needs to be selected manually and this leads to data-sets that reflect the supervisor preferences.
Structured clutter represents the second main aspect of the {iCubWorld} data-set. In image retrieval settings images are selected from very different contexts, the background is highly variable, with the exception of some useful contextual information \cite{Tor2003}. On the contrary, in our setting images are acquired in a much more limited environment (a robotics lab) and displayed a much more stable background. 
On these respects, the iCubWorld data-set appears to be different from the existing benchmarks.

%% file: experiments.tex
\section{Methods}
We briefly review the current state of the art algorithms for visual recognition tasks, with an emphasis on the so-called hierarchical models \cite{Lazebnik06,yang09, boureau11,serre07} which have been proved effective on the existing image retrieval benchmarks. 

Typically, an unsupervised learning stage is employed to obtain representations that take into account the domain of the problem considered. These representations are encoded in compact vectors of fixed dimension, that are fed to a classifier in order to learn the visual appearance of individual categories.
More specifically, a sequence of descriptors 
is extracted to encode the local responses of the image with respect to a predefined (or in some cases learned from data) set of filters. Common filters are image patches~\cite{feifei05}, SIFT~\cite{lowe04}, SURF~\cite{bay08} or Gabor Filters \cite{serre07}. In categorization problems descriptors are extracted from a dense regular grid on the image, following the study in~\cite{feifei05}. 

Higher level representations are then built on top of local descriptors. 
In general, an (unsupervised) learning step adapts the representation to the data. 
Often a set or dictionary of atoms is learned from data, and subsequently used to code the available images. Well known methods for learning the dictionaries are $K$-Means \cite{csurka04}, or Dictionary Learning techniques \cite{lee07, yang09}. Examples of coding methods are Vector Quantization (VQ) \cite{Lazebnik06}, Sparse Coding (SC) \cite{yang09} and Locality-constrained Linear Coding (LLC) \cite{wang10}. 
The coding stage produces again a set of local coded descriptors, 
then a pooling map combines them and encodes higher-level statistics of the image. It has been empirically observed instead, that Sparse Coding (SC) favors max pooling over average pooling \cite{boureau10}.
This procedure is often associated to a spatial pyramid representation \cite{Lazebnik06} of the image. In this case coding is applied to overlapping regions of the image pyramid and then all descriptions are simply concatenated.  

A different perspective is given by the HMAX biologically inspired framework \cite{serre07}, which is an algorithmic model of the recognition process in humans. HMAX retraces the humans ventral stream structure of simple and complex cells forming a hierarchy of alternating layers (see \cite{serre07} for more details). However the underlying implementation, alternating filtering and pooling stages,  has analogies with standard visual recognition systems.

\begin{figure}[t]
\begin{center}
   \includegraphics[width=\columnwidth]{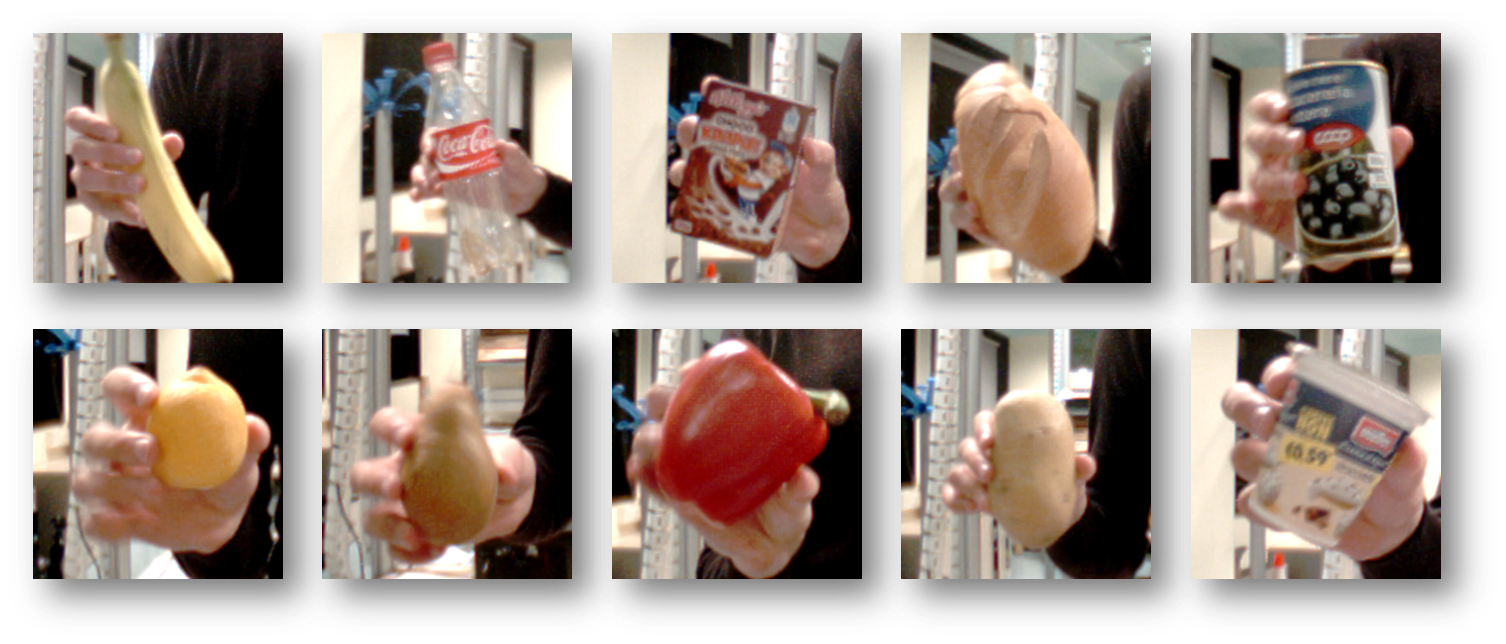}
\caption{The iCubWorld data-set. $10$ categories: Bananas, Bottles, Boxes, Bread, Cans, Lemons, Pears, Peppers, Potatoes, Yogurts. Each category contains $4$ different instances.}
\label{fig:dataset}
\end{center}
\end{figure}
\section{Evaluation}
%
We evaluate the iCubWorld data-set with a set of prototypical methods from the literature:
\begin{itemize}
\item {\bf BOW} or Bag of Words ~\cite{csurka04} consisting of a K-means followed by vector quantization and average pooling.
\item {\bf SC} or Sparse Coding~\cite{yang09}, including a dictionary learning step, followed by sparse coding and a max pooling.
\item{\bf LLC} or Locality-constrained Linear Coding \cite{wang10}.
\item {\bf HMAX} ~\cite{serre07}.
\end{itemize}
Besides HMAX, which is based on Gabor filters and has a slightly different structure, the other methods share the same SIFT~\cite{lowe04} feature extraction stage and are organized in a three-levels Spatial Pyramid Representation~\cite{Lazebnik06}. SIFT features are extracted from a dense grid on the image with granularity of $8$ pixels on patches of $16 \times 16$ pixels. The dictionary size is set to $1024$.
Instead, since HMAX does not employ a pyramidal pooling, we employ a dictionary of $4096$ features for a fair comparison. 
An SVM~\cite{vapnik98} classifier is used to train and test the system. 

We perform five different experiments to emphasize the main aspects of the iCubWorld data-set. For all the experiments, the system has been trained on the same set of data (described in Sec.~\ref{sec:dataset}) of $10$ categories, $3$ object instances per category and $200$ examples per instance ($600$ images per category).
%
%
%
Fig. \ref{fig:test} reports, for one object class,  a sample frame of the 3 object instances used for training (first row) together with samples of each of the $5$ test sets.

The experimental analysis is based on a frame-based accuracy, with the exception of Test $4$ where we also tried to exploit the whole video sequence.

\subsection{Test $T1$: Known instances}

This experiment is a sanity check to verify the consistency of the proposed dataset: we test the system on a new set of images depicting the same object instances used for training  shown by the same supervisor. Results in Tab.~\ref{tab:benchmark} (first column) report the mean accuracy over the $10$ categories, showing that all methods lead to a good representation of the observed data.

\subsection{Test $T2$: Generalizing w.r.t. the supervisor}
\label{sec:t2}
In this test the human supervisor changes with respect to the one that took care of the training section. We test the system \ on the same object instances used for training. In this case we experience a remarkable performance drop (see Tab. \ref{tab:benchmark}, second column). The best performing image representation is SC with an overall mean accuracy of $38.2\%$. We argue that this dramatic drop is due to the different appearance of the supervisor and the fact he may possibly move the object in a different way: thus in the test sequences the object may be shown w.r.t. new points of view and within a different background.
\begin{figure}[t]
\begin{center}
   \includegraphics[width=\columnwidth]{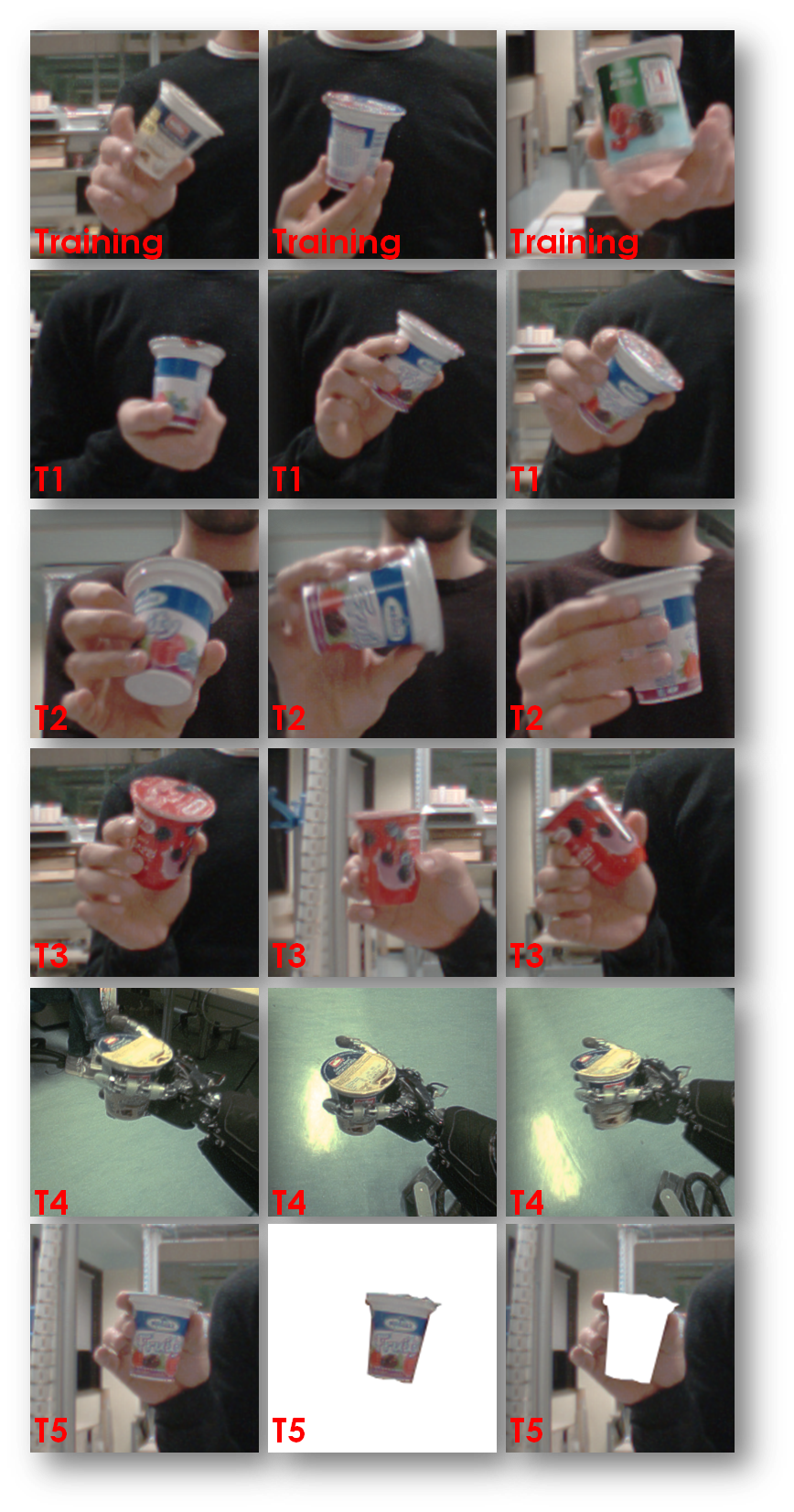}
\caption{Samples of training and test sets for the object class ``yogurt". }
\label{fig:test}
\end{center}
\end{figure}

\subsection{Test $T3$: Generalizing w.r.t. the object category}
\begin{figure}[t]
\begin{center}
   \includegraphics[width=\columnwidth]{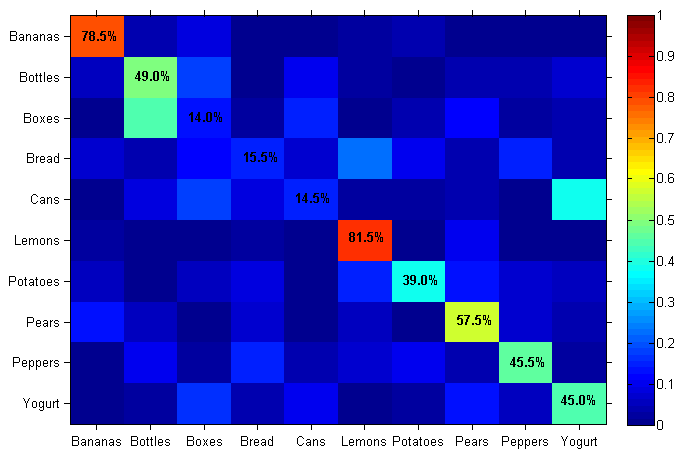}
\caption{Confusion matrix obtained with SC applied to the test set T3.}
\label{fig:cm}
\end{center}
\end{figure}

The third test is aligned with standard computer vision datasets, where the goal is to recognize new instances of the same object category. For this test we use the same supervisor of the training phase, while we choose an unknown instance of the known object categories. Again this test leads to low performances:  the highest score is obtained by SC ($44.0\%$); comparing it to results obtained with the same method to other datasets such as Caltech-101, we have a drop of $30\%$ of the accuracy even though only $10$ categories are used. The reached accuracy is comparable to the one obtained by SC on Caltech-256 ($40.14\%$ see \cite{yang09}), although the iCubWorld appears to be simpler. 
Fig. \ref{fig:cm} reports the confusion matrix obtained in this case by SC. The appearance of most categories is not correctly captured by the method.
\subsubsection{The Impact of the Dictionary Size}
\label{sec:dictionary}
\begin{figure}[t]
\begin{center}
   \includegraphics[width=\columnwidth]{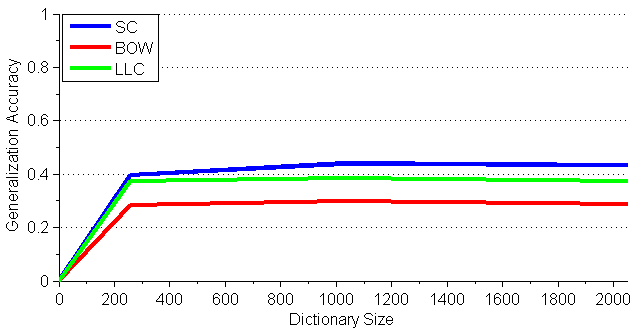}
\caption{The generalization performances with respect to the dictionary size. Highest accuracy is obtained then the number of atoms is 1024.}
\label{fig:dict}
\end{center}
\end{figure}

Given the results obtained in the previous test, we tried to investigate the impact of different dictionary sizes. In Fig. \ref{fig:dict}, we show the mean accuracy in categorization for sizes ranging from $K=256$ to $K=2048$. Performances seem to be not affected by the number of atoms and the highest accuracy is reached for $K=1024$. This means that the number of bases is already able to catch the class variability, therefore the complexity of the problem lies in the chosen setting.

\subsection{Test $T4$: Changing the acquisition domain}
\label{sec:t4}
What we would expect from a visual recognition system in a robotics setting is the capability to generalize even if we try to change the domain. For this setting we used sequences of objects acquired in the Robot Mode (Sec. \ref{sec:dataset}), letting the robot grasp the objects (see Fig. \ref{fig:test}, fourth row). One video for each category has been acquired. Even in this case poor performances are obtained (see Tab. \ref{tab:benchmark}, last column). In this case we also tried to carry out a video-based analysis of the results, with a simple winner-takes-all voting scheme. Again, the highest accuracy has been obtained by SC ($20.0\%$) that correctly classified only two categories: bottles and peppers.

\subsection{Test $T5$: Are we really learning the objects?}
\label{sec:structured_clutter}
After the  tests described above, we started to investigate the possible causes of the low performances obtained. The reason appears to be the presence of structured clutter that cannot be used as further context to learn the objects. In other words, our conjecture is that current visual recognition systems work well only when the context can be used profitably. To confirm this hypothesis, we selected a subset of $10$ images per category, where the classifiers perform an overall mean accuracy of $99.0\%$. We manually segmented the object obtaining two binary masks: one for the object and one for the background (see Fig. \ref{fig:test}, last row). Thereafter we try to classify both the images where only the segmented object is presented and images where only the background is shown. 
A first surprising result is that the background is already enough to obtain $68.0\%$ of mean accuracy, even if the object is not present in the scene (see Tab. \ref{tab:bkg}). On the other hand, classification performed on the segmented objects is $80.0\%$. This means that the context contributes for $20.0\%$ of accuracy and confirms our intuition that both object and context have been learned.

\begin{table}[t]
\begin{center}
\rowcolors{2}{gray!35}{}
\begin{tabular}{lcccc}
	
	&  T1(\%)	 &  T2(\%) & T3(\%)   & T4(\%)  \\

\toprule

 BOW & 78.6 & 27.8 & 29.8 & 14.4  \tstrut \bstrut  \\

 SC &   89.8  & {\bf 38.2} & {\bf 44.0} & {\bf 19.2}  \tstrut \bstrut  \\

 LLC  &  87.8  & 35.7 & 38.4  & 13.5 \tstrut \bstrut  \\

 HMAX  & {\bf 91.6}  & 36.0 & 41.9 & 17.2 \tstrut \bstrut  \\
\end{tabular}
\end{center}
\caption{
\footnotesize{Classification accuracy averaged over $10$ categories. 600 training examples per category. Each column represents a particular test set.}
}
\label{tab:benchmark}
\end{table}

\begin{table}[t]
\begin{center}
\rowcolors{2}{gray!35}{}
\begin{tabular}{lccc}
	\textbf{T5}	&  Whole(\%)	 &  Obj(\%) & Bkg(\%)    \\
\toprule
SC  &   99.0  & 80.0 & 68.0   \tstrut \bstrut  \\
\end{tabular}
\end{center}
\caption{
\footnotesize{Test $T5$ results. \textit{Whole}: the entire bounding box has been used for object modeling. \textit{Obj}: only the segmentation around the object is used for feature extraction. \textit{Bkg}: features have been extracted only from the background.}
}
\label{tab:bkg}
\end{table}

%% file: discussion.tex
\section{Discussion}
In this paper we presented the iCubWorld data-set, acquired with the help of a HRI scheme. The data-set  currently includes 10 object categories, but is meant to grow in the near future, thanks to the simplicity of data acquisition and annotation. 
We analysed the performances of a selection of visual recognition methods from the state-of-the-art over the iCub data-set. The results we achieved confirm
the complexity of the data-set, comparable to standard computer vision benchmarks, and the presence of new challenges with respect to image retrieval data-sets. In particular, a structural background carrying little or no context information for the object, highlighted some limits of BOW-like methods. 
Hopefully, having the iCubWorld data-set publicly available, the computer vision research community could find new insights and motivations to develop systems able to work effectively also in the Human Robot Interaction applications.